%% file: main.tex
\def\year{2021}\relax
\documentclass[letterpaper]{article} 
\usepackage{aaai21}  
\usepackage{times}  
\usepackage{helvet} 
\usepackage{courier}  
\usepackage[hyphens]{url}  
\usepackage{graphicx} 
\usepackage{natbib}  
\usepackage{caption} 
\usepackage{amsfonts}
\usepackage{amsthm}

\newtheorem{mydef}{Definition}

\newtheorem{corollary}{Corollary}

\newtheorem{lemma}{Lemma}

\newtheorem{proposition}{Proposition}
\newtheorem{remark}{Remark}

\title{The Effect of Prior Lipschitz Continuity on the Adversarial Robustness of Bayesian Neural Networks}

\author{
    Arno Blaas \hspace{0.5cm} 
    Stephen J. Roberts 
    \\
}
\affiliations{
    Machine Learning Research Group \&  Oxford-Man Institute\\
    University of Oxford\\
    \{arno,sjrob\}@robots.ox.ac.uk
}

\begin{document}

\maketitle

\begin{abstract}
It is desirable, and often a necessity, for machine learning models to be robust against adversarial attacks. This is particularly true for Bayesian models, as they are well-suited for safety-critical applications, in which adversarial attacks can have catastrophic outcomes. In this work, we take a deeper look at the adversarial robustness of Bayesian Neural Networks (BNNs). In particular, we consider whether the adversarial robustness of a BNN can be increased by model choices, particularly the Lipschitz continuity induced by the prior. Conducting in-depth analysis on the case of i.i.d.,  zero-mean Gaussian priors and posteriors approximated via mean-field variational inference, we find evidence that adversarial robustness is indeed sensitive to the prior variance.
\end{abstract}


\section{Introduction}
\input{sections/introduction.tex}

\section{Background}
\input{sections/background.tex}

\section{The Adversarial Robustness of Bayesian Neural Networks}
\input{sections/method.tex}

\section{Experiments}
\input{sections/experiments.tex}

\section{Discussion}
\input{sections/discussion.tex}

\section{Acknowledgments}
We thank Samuel Kessler for helping with the implementation of the MFVI training. AB is supported by the Konrad-Adenauer-Stiftung.

\bibliography{bibfile.bib}

\section{Appendix}
\input{sections/appendix.tex}

\end{document}

%% file: sections/introduction.tex
Adversarial attacks (i.e.\ input points intentionally crafted to trick a model into misclassification) have raised serious concerns about the safety and security of models learned from data \citep{biggio2018wild}. In the case of safety-critical applications, such as healthcare \citep{finlayson2018adversarial} or autonomous driving \citep{mcallister2017concrete}, the impact of adversarial attacks could be catastrophic. For such applications, Bayesian machine learning models, such as Bayesian Neural Networks (BNNs) are a natural choice, as they allow the incorporation of predictive uncertainty into decision making \citep{fan2020bayesian}. While the robustness of models against adversarial attacks (adversarial robustness) has been extensively studied for non-Bayesian neural networks \citep{katz2017reluplex,huang2017safety,biggio2018wild, yang2019me}, the robustness of BNNs has been little understood so far. 
In particular, it remains unclear which model choices can be made to increase the adversarial robustness of Bayesian models. Drawing on the established link between Lipschitz continuity and adversarial robustness \cite{yang2020closer, szegedy2013intriguing}, we investigate this question by analysing the effect the Lipschitz continuity, induced by the prior distribution over neural network parameters, has on the adversarial robustness of the resulting BNN. This results in the following contributions.

\subsection{Contributions}
The specific question we address in this work is:
\begin{itemize}
    \item Can the adversarial robustness of BNNs be increased by specific model choices concerning the Lipschitz continuity induced by prior distributions? 
\end{itemize}
We focus on the case of i.i.d. Gaussian priors over neural network parameters and posteriors approximated by mean-field variational inference (MFVI).\\ 
Firstly, we show both in theory and empirically, that a lower prior Lipschitz constant, induced by a lower prior variance, decreases the posterior global Lipschitz constant of a BNN.
Secondly, we investigate empirically if this decreased posterior Lipschitz constant leads to increased adversarial robustness. Our findings are mixed. While the adversarial loss does generally decrease as the posterior Lipschitz constant decreases, the adversarial accuracy does not necessarily increase.

\subsection{Related Work}
Despite being well-suited for safety-critical applications \citep{mobiny2019risk}, the question of the adversarial robustness of BNNs and Bayesian models in general has been the subject of relatively few works compared to the literature for deterministic neural networks (DNNs). There are some papers that derive probabilistic or statistical adversarial robustness guarantees for BNNs \cite{cardelli2019statistical, wicker2020probabilistic} and other Bayesian models such as Gaussian process classification models \cite{blaas2020adversarial}. 
Yet, unlike our work, these approaches do not analyse ways to increase the adversarial robustness of a BNN. 
Closest to this approach is the work of \cite{liu2018adv}, who establish that adversarial training can increase the adversarial robustness of mean-field variational inference approximated BNNs. Our work differs from their approach in that we look at approaches to increase the adversarial robustness of BNNs through model choices, rather than via data manipulation (adversarial training in effect being a form of data augmentation).

%% file: sections/background.tex
\subsection{Bayesian Neural Networks}

Let $\mathcal{D}_n$ be a training data set consisting of tuples $(x_i,y_i)_{\{1\leq i \leq n}$.
Unlike DNNs, which work on the assumption that a single parameter configuration, namely the one which best explains $\mathcal{D}_n$, should be used for prediction, Bayesian neural networks (BNNs) \cite{Denker1991, MacKay1992} use a weighted set of parameter configurations, which all explain $\mathcal{D}_n$ to different extents. This provides predictions as well as associated uncertainty over function behaviour, with uncertainty increasing naturally in unobserved parts of the domain. The weighting of parameter configurations is performed using Bayesian inference, namely specifying a prior distribution, $p_0(\theta)$, over the parameters of a neural network, as well as defining likelihood of the data under the model, $p(\mathcal{D}_n | \theta)$. For classification, categorical distributions with class probabilities given by the neural network (NN) output are assumed, i.e.
\begin{equation} \label{eq:BNNlikelihood1}
p(\mathcal{D}_n | \theta) = \prod_{i=1}^n {p(y_i|x_i,\theta)},
\end{equation} 
where 
\begin{equation} \label{eq:BNNlikelihood2}
p(y_i|x_i,\theta) = f^{c=y_i}(x_i|\theta)
\end{equation} 
with $f^{c}$ being the NN output in dimension $c$.
Applying Bayes' formula allows us to derive the posterior distribution over the parameters $\theta$:
\begin{equation} \label{eq:BNNposterior}
p(\theta | \mathcal{D}_n) = \frac{\prod_{i=1}^n {f^{c=y_i}(x_i|\theta)}p_0(\theta)}{\int{\prod_{i=1}^n {f^{c=y_i}(x_i|\theta)}p_0(\theta)}d\theta}.
\end{equation}
For prediction, the parameter configurations, $\theta$, are then weighted by this posterior distribution:
\begin{equation} \label{eq:BNNpred}
p(y_* = c | x_*, \mathcal{D}_n) = \int{ f^{c}(x_*|\theta) p(\theta | \mathcal{D}_n)}d\theta. 
\end{equation}

\paragraph{Approximating the posterior:} Unfortunately, the posterior distribution $p(\theta | \mathcal{D}_n)$ in Eq. (\ref{eq:BNNposterior}) is typically analytically intractable for NNs and thus needs to be approximated. 
For this work, we focus on the commonly used mean-field variational inference (MFVI) approximation method \cite{blundell2015weight}.
MFVI approximates $p(\theta | \mathcal{D}_n)$ with a simple multivariate Gaussian distribution $q_{\mathbf{m},\mathbf{s}}(\theta) = \mathcal{N}(\theta | \mathbf{m},\mathrm{diag}(\mathbf{s})I)$ by maximising the evidence lower bound
\begin{eqnarray} \label{ELBO}
\mathrm{ELBO}(\mathbf{m},\mathbf{s}) & = &\int \prod_{i=1}^n {f^{c=y_i}(x_i|\theta)} q_{\mathbf{m},\mathbf{s}}(\theta) d\theta  \nonumber \\
&& -  \mathrm{KL}(q_{\mathbf{m},\mathbf{s}}(\theta) || p_0(\theta)),
\end{eqnarray}
where $\mathrm{KL}q_{\mathbf{m},\mathbf{s}}(\theta) || p_0(\theta))$ is the KL-divergence between $q_{\mathbf{m},\mathbf{s}}(\theta)$ and the prior $p_0(\theta)$. 
The ELBO (Evidence Lower Bound) allows the equivalence between maximising $\log p(\mathcal{D}_n) - \mathrm{KL}(q_{\mathbf{m},\mathbf{s}}(\theta) || p(\theta | \mathcal{D}_n))$, (i.e. simultaneously maximising log-evidence) and minimising the KL-divergence between $q_{\mathbf{m},\mathbf{s}}(\theta)$ and the posterior $p(\theta | \mathcal{D}_n)$. 
Despite the seemingly simple structure of the functional $q$, it can be shown that MFVI can have universal approximation properties \cite{farquhar2020liberty}.

\subsection{Adversarial Robustness}

Adversarial robustness can be defined in multiple ways, particularly for Bayesian models.  Definitions include detection mechanisms based on model outputs, such as predictive uncertainty \cite{smith2018understanding} and density estimators \cite{feinman2017detecting}. Since, however, it has been demonstrated that including such artifacts into the attack objective can circumvent these mechanisms \cite{grosse2018limitations}, we choose to restrict our analysis to the traditional definition of adversarial robustness \cite{croce2020provable}. 
\begin{mydef}[Adversarial robustness] \label{def:rob}
The robustness of a BNN classifier at point $x$ belonging to class $c$ with respect to an $L_p-$norm $\|\cdot\|$ is defined as 
\begin{eqnarray} \label{eq:robustness}
    &&\mathbf{r}_p(x) = \min_{\delta} \|\delta\|  \\
    &&\mathrm{s.th. } ~ \max_{b\neq c} p(y = b|x+\delta, \mathcal{D}_n) \geq p(y = c|x+\delta, \mathcal{D}_n) \nonumber
\end{eqnarray}
\end{mydef}
The general adversarial robustness of a classifier for a given data set is then usually classified as the minimum or mean adversarial robustness across all test points. 

\subsection{Lipschitz Continuity and Adversarial Robustness}

Multiple works have established the model independent link between Lipschitz continuity and adversarial robustness \citep{szegedy2013intriguing, hein2017formal, yang2020closer}. We here briefly review the concept of Lipschitz continuity and outline its relation to adversarial robustness.

\begin{mydef}[Lipschitz continuity]
A function $g: \mathcal{X} \to \mathcal{Y}$ is called (globally) Lipschitz continuous with respect to metrics $\mathfrak{d}_{{X}}: \mathcal{X} \times \mathcal{X} \to \mathbb{R}_{\geq 0}$ and $\mathfrak{d}_{Y}: \mathcal{Y} \times \mathcal{Y} \to \mathbb{R}_{\geq 0}$ if there exists a constant $L \in \mathbb{R}_{\geq 0}$ for which
\begin{equation} \label{eq:LipCont}
\forall x,x' \in \mathcal{X}: ~~ \mathfrak{d}_{Y}(g(x),g(x')) \leq L~ \mathfrak{d}_{X}(x,x')
\end{equation}
\end{mydef}

A constant $L \in \mathbb{R}_{\geq 0}$ that fulfils Eq. (\ref{eq:LipCont}) is called a (global) Lipschitz constant of $g$, and the smallest such constant is called the best (global) Lipschitz constant of $g$, denoted by $L^*(g)$.
If Eq. (\ref{eq:LipCont}) only holds for a subset $U \subset \mathcal{X}$, we say that $g$ is locally Lipschitz continuous with local Lipschitz constant $L$ in $U$. 
In the context of adversarial robustness we are in particular interested into $L_p-$norm induced metrics of the form $\mathfrak{d}_{{X/Y}}(\eta,\eta ') = \|\eta-\eta '\|$ for some $L_p-$norm $\|\cdot\|$.
A small (global or local) Lipschitz constant $L$ of a model prediction function $g$ hence implies high adversarial robustness, since from $\|\delta\| \leq \gamma$ it follows $\|g(x+\delta) - g(x) \| \leq L \gamma$. 

%% file: sections/method.tex
The adversarial robustness of a BNN is entirely determined by its predictive distribution $p(y_* = c | x_*, \mathcal{D}_n)$ in Eq. (\ref{eq:BNNpred}). It is thus a combination of the functional form of $f^c(\cdot)$ (the NN architecture) and the posterior $p(\theta|\mathcal{D}_n)$. From Eq. (\ref{eq:BNNposterior}) we see that the posterior, in turn, is entirely determined by the NN architecture, the observed data $\mathcal{D}_n$ and the prior distribution, $p_0(\theta)$. Assuming the data $\mathcal{D}_n$ to be given and excluding its manipulation from the inference procedures under inspection, this leaves two components that determine the adversarial robustness of a BNN: the NN architecture and the prior distribution $p_0(\theta)$. 
The analysis of the impact of NN architecture is important, but not the main focus of our work, as its impact is identical for DNNs and BNNs and has already been extensively researched for DNNs \cite{guo2020meets, hendrycks2019benchmarking, xie2019feature, szegedy2013intriguing, cubuk2017intriguing}. 
We thus primarily focus on the impact of $p_0(\theta)$ on adversarial robustness. We consider here if certain priors on the parameters induce higher adversarial robustness of the posterior prediction function $p(y_* = c | x_*, \mathcal{D}_n)$. To simplify the analysis, we restrict the NN architectures under consideration to be a fully connected feed-forward NNs of the form:
\begin{eqnarray}
    f^c(x|\theta) = \mathrm{sm}_c(W^D\phi (W^{D-1} \phi (\ldots \phi (W^1x + b^1)  \nonumber\\
    \ldots  )+ b^{D-1})+ b^D)  \label{eq:FFnet}
\end{eqnarray}
where $\mathrm{sm}$ is the softmax function, $\phi$ the activation function (e.g. ReLu) and $\theta = (W^D, W^{D-1}, \ldots W^1, b^D, b^{D-1} \ldots, b^1)$. We write $f^c_{\theta}$ for $f^c(\cdot|\theta)$.

\subsection{Effect of the Prior on Adversarial Robustness: Lipschitz Continuity by Prior Variance Reduction}

Given the link between Lipschitz continuity and adversarial robustness, we focus on the effect of inducing Lipschitz continuity properties, on functions drawn from the prior, on the adversarial robustness of the prediction function $p(y_* = c | x_*, \mathcal{D}_n)$. 
As a starting point, we consider the most commonly used tractable prior $p_0(\theta) = \mathcal{N}(\theta|0, \alpha^2 I)$ \cite{wilson2020bayesian, wenzel2020good, Gal2016}. For a fixed NN architecture, the Lipschitz continuity of $f^{c}(\cdot|\theta)$ for $\theta \sim p_0(\theta)$ is then only controlled by the prior variance scaling factor, $\alpha^2$. The following two results show that the relationship between $\alpha^2$ and the Lipschitz continuity of $\int{f^{c}(\cdot|\theta) p_0(\theta)}d\theta$ is monotonic. Proofs can be found in the Appendix.
\vspace{0.1cm}

\begin{proposition} \label{prop:lipBNN}
Let $f^{c}_{\theta}$ be a feed-forward NN as defined in Eq. (\ref{eq:FFnet}), with activation functions $\phi$ that are $k$-Lipschitz continuous\footnote{For example, for ReLu or tanh activation functions $k=1$.}. For $g(x) = \int{ f^{c}(x|\theta) p(\theta)}d\theta$ with some probability density function $p(\theta)$ it holds that:
\begin{equation}
|g(x) - g(x')| \leq k^D \prod_{l=1}^D \mathbb{E}_{p}[\|W^l\|] \|x - x'\|,
\end{equation}
i.e. $g$ is Lipschitz continuous with constant $k^D \prod_{l=1}^D \mathbb{E}_{p}[\|W^l\|]$.  
\end{proposition}
\vspace{0.1cm}

\begin{proposition} \label{prop:lipNormalBNN}
For $p(\theta)$ with $W^l \sim \mathcal{N}(W^l|M^l,\alpha^2 I)$, the Lipschitz constant of 
$g$ in Proposition \ref{prop:lipBNN} decreases monotonically with $\alpha^2$ and with $\|M^l\|$.
\end{proposition}
\vspace{0.1cm}

\begin{remark}
Note that the last equation implies that for $\alpha^2 \to 0$, $L(g) \to \prod_{l=1}^D \|M^l\|$.
\end{remark}
\vspace{0.1cm}

We thus conclude that for the tractable prior $p_0(\theta) = \mathcal{N}(\theta|0,\alpha^2 I)$, we can analyse the effect of varying Lipschitz continuity inducing prior distributions on adversarial robustness of the posterior predictive model by analysing the effect of the prior variance $\alpha^2$, as smaller $\alpha^2$ implies smaller Lipschitz constants of the resulting mean prior function $g(x) = \int{ f^{c}(x|\theta) p(\theta)}d\theta$.\\
Intuitively, properties of the prior transfer to the posterior and we would thus expect that a smaller variance $\alpha^2$ of the prior also implies a smaller Lipschitz constant of the posterior which in turn implies higher adversarial robustness.
If the posterior is inferred using MFVI, this can actually be shown theoretically.
\begin{lemma} \label{lem:priorposterior}
Let $p_0(\theta) = \mathcal{N}(\theta|0, \alpha^2 I)$ and $f^c_{\theta}$ defined as in Eq. (\ref{eq:FFnet}). For $p(\theta | \mathcal{D}_n) \approx q_{\mathbf{m},\mathbf{s}}(\theta) = \mathcal{N}(\theta | \mathbf{m},\mathrm{diag}(\mathbf{s})I)$ obtained by maximising Eq. (\ref{ELBO}), we have that both $\|\mathbf{m}\|_2$ as well as all elements of $\mathbf{s}$ are monotonic functions of $\alpha^2$. 
\end{lemma}

\begin{corollary} \label{cor:MFVI}
For $p_0(\theta)$ and $f^c_{\theta}$ defined as above, the Lipschitz constant defined in Proposition \ref{prop:lipBNN} of BNNs with MFVI approximated posterior distribution is a monotonic function of the prior variance $\alpha^2$.
\end{corollary}

For MFVI approximated posteriors with i.i.d. Gaussian priors $p_0(\theta) = \mathcal{N}(\theta|0, \alpha^2 I)$, we thus conclude that a decrease in prior variance $\alpha^2$ decreases the Lipschitz constant of the resulting BNN. As a low Lipschitz constant is a sufficient condition for high adversarial robustness (\cite{szegedy2013intriguing}, of course at the price of potentially lower accuracy), lowering the prior variance $\alpha^2$ should thus increase adversarial robustness. However, there is one crucial condition that this conclusion relies on - namely that the Lipschitz constant derived above is reasonably tight. While it is guaranteed to converge to $0$ if $\alpha^2 \to 0$, we don't have a result that describes the rate of convergence. If infinitesimal values of $\alpha^2$ are required in practice to obtain fairly low values of the Lipschitz constant of the posterior, the results above only tell us the obvious result that falling back to a constant predictor with essentially $0$ mean and $0$ variance results in low Lipschitz continuity. In order to see how the established relationships behave in practice, we thus conduct some experiments in the next section.\\


%% file: sections/experiments.tex
We conduct experiments on two datasets:  MNIST \cite{lecun1998mnist}, and FashionMNIST \cite{xiao2017/online}. All inputs are scaled to $[0,1]$. For both data sets, we limit the training data set size to $5'000$ and the test data set size to $1'000$ for computational reasons. We set $f^c_\theta$ to be a small feed-forward neural network with ReLu activation functions and three hidden layers of $[32,64,32]$ neurons respectively. For each experiment, we use MFVI to approximate the BNN that results from combining $f^c_\theta$ with Gaussian i.i.d. priors $p_0(\theta) = \mathcal{N}(\theta|0,\alpha^2 I)$ and the data. To this end, we maximise the ELBO (Eq. (\ref{ELBO})) using SGD for $100$ epochs. The BNN prediction function is then approximated using the Monte Carlo (MC) estimate $p(y_* = c | x_*, \mathcal{D}_n) \approx \frac{1}{T}\sum_{t=1}^T{ f^{c}(x_*|\theta_t)}$ with $\theta \sim q(\theta | \mathbf{\bar{m}}, \mathrm{diag}(\mathbf{\bar{s}}) I)$ using $T=100$ samples.\\
By varying the values $\alpha^2$, we inspect the effect of the Lipschitz continuity of $p_0$ on the adversarial robustness of the BNN. We measure adversarial robustness empirically using three different variants of PGD with $20$ iterations (see Appendix). The perturbation budget is set to $0.1$ in the $L_{\infty}-$norm.\\
Also, we calculate an estimate of the global Lipschitz constants as on the test data set $\mathcal{T}_m$ as
\begin{equation} \label{eq:lipEst}
    L(g) = \max_{x_i, x_j \in \mathcal{T}_m} \frac{\|g(x_i) - g(x_j)\|_{p}}{\|x_i - x_j\|_{p}}
\end{equation}
which is a lower bound to to the true global Lipschitz constant \cite{blaas2019localised}.
We analyse Lipschitz continuity with respect to both the $L_2-$ and the $L_{\infty}-$norm as they have been described as corresponding the most to human perception \cite{madry2017towards, goodfellow2014explaining}.

\subsection{Experiment 1: MNIST}

\begin{table}[]
    \centering
    \begin{tabular}{r||c|c|c|c}
    & \multicolumn{4}{c}{Prior variance $\alpha^2$}\\
        &  0.01 & 0.1 & 1.0 & 10.0  \\
        \hline 
        Training Acc.  & 11.3  & 99.7   & 100.0  & 99.1 \\
        Clean Test Acc.  & 10  & 94.5   & 94.8  & 93.3 \\
        $L_2$ prior Lip. const.   & 0.0 & 0.9   & 126.2  & 13416.7   \\
        $L_{\infty}$ prior Lip. const.   & 0.0  & 4.8   & 591.1    & 60513.4 \\
        $L_2$ post. Lip. const.   & 0.0 & 3.7   & 7.6  & 32.6   \\
        $L_{\infty}$ post. Lip. const.   & 0.0  & 28.4   & 55.8    & 224.7    \\
        Adv. Test Loss   & 2.3  & 3.6   & 6.6    & 18.5    \\
        Adv. Test Acc. & 10.0  & 23.6  &  30.9  & 20.5 \\
        Net Adv. Test Acc. & 100.0  & 28.3  &  35.7  & 25.4   \\
        
    \end{tabular}
    \caption{Effect of prior Lipschitz continuity (as controlled by prior variance $\alpha^2$) on posterior adversarial robustness for MNIST. Lipschitz constants $L(g)$ (Eq. (\ref{eq:lipEst})) are calculated on logits, i.e. pre-softmax, to make differences clearer. Net Adv. Test Acc. compares predicted class of attack with predicted class of original test point rather than with correct class (see Definition \ref{def:rob}). Values show averages over five random seeds $\{10,20,30,40,50\}$}
    \label{tab:VIMNIST}
\end{table}
The results are shown in Table \ref{tab:VIMNIST}. It can be seen that a lower prior variance, $\alpha^2$, corresponding to a lower prior Lipschitz constant, indeed leads to a lower posterior Lipschitz constant. As hoped, this does correspond to a higher adversarial robustness when measured by the cross-entropy loss of the perturbed images ('Adv. Test Loss'). Also, as predicted by the theory, for small enough $\alpha^2$, the BNN stays a constant function which naturally has the highest net adversarial test accuracy (at the price of lowest clean accuracy). 
However, while the adversarial test loss is monotonic in $\alpha^2$, this is not true for the (net) adversarial test accuracy, as $\alpha^2 = 1.0$ yields better (net) adversarial accuracy than $\alpha^2 = 0.1$.


\subsection{Experiment 2: FashionMNIST}

\begin{table}[]
    \centering
    \begin{tabular}{r||c|c|c|c}
    & \multicolumn{4}{c}{Prior variance $\alpha^2$}\\
        &  0.01 & 0.1 & 1.0 & 10.0  \\
        \hline 
        Training Acc.  & 10.5  & 91.2   & 94.6  & 87.9 \\
        Clean Test Acc.  & 8.8  & 83.4   & 82.7  & 80.5 \\
        $L_2$ prior Lip. const.   & 0.0 & 1.7   & 203.3  & 20818.4   \\
        $L_{\infty}$ prior Lip. const.   & 0.0  & 8.8   & 1004.6    & 102553.9 \\
        $L_2$ post. Lip. const.   & 0.0 & 3.6   & 12.7  & 32.3   \\
        $L_{\infty}$ post. Lip. const.   & 0.0  & 26.8   & 75.3    & 177.4    \\
        Adv. Test Loss   & 2.3  & 6.9   & 19.0    & 17.5    \\
        Adv. Test Acc.   & 8.8  &  6.2  & 8.3  & 8.8   \\
        Net Adv. Test Acc.   & 100.0 & 16.0   & 17.1  & 18.2   \\
        
    \end{tabular}
    \caption{Effect of prior Lipschitz continuity (as controlled by prior variance $\alpha^2$) on posterior adversarial robustness for FashionMNIST. Lipschitz constants $L(g)$ (Eq. (\ref{eq:lipEst})) are calculated on logits, i.e. pre-softmax, to make differences clearer. Net Adv. Test Acc. compares predicted class of attack with predicted class of original test point rather than with correct class (see Definition \ref{def:rob}). Values show averages over five random seeds $\{10,20,30,40,50\}$.}
    \label{tab:VIFashionMNIST}
\end{table}

The results are shown in Table \ref{tab:VIFashionMNIST}.
Again, we see that as theory predicts, a lower prior variance $\alpha^2$ corresponding to a lower prior Lipschitz constant indeed leads to a lower posterior Lipschitz constant. However,for FashionMNIST, this does not necessarily translate to a higher adversarial robustness even when adversarial test loss is considered, as the test loss for $\alpha^2 = 10$ is slightly lower than the one for $\alpha^2 = 1$. In terms of net adversarial robustness, the relationship to prior variance is the inverse of what one would hope for, as larger prior variance (beyond $0.1$) leads to higher net adversarial accuracy, even though the increase is only marginal. 

%% file: sections/discussion.tex
We here investigate the question as to whether the adversarial robustness of BNNs can be increased by specific model choices concerning the Lipschitz continuity induced by prior distributions.
For the case of i.i.d. zero-mean Gaussian priors, we have shown that a lower prior Lipschitz constant, induced by a lower prior variance $\alpha^2$, reduces the posterior Lipschitz constant of a BNN whose posterior is approximated using MFVI. The desired increase in adversarial robustness was, however, only partially observed. This could be due to the fact that the obtained adversarial robustness is empirical, i.e. the values presented give an upper bound to the theoretical adversarial accuracy over the test set. Another, more interesting explanation is that the adversarial robustness is not driven by the Lipschitz constant alone, but also by the margin (i.e. difference between largest and second largest class outputs $f^c$). Potentially, the effect of the prior variance has an effect on the margin which offsets the increased Lipschitz constant. However, the fact that the relationship between Lipschitz constant and adversarial test loss is mostly monotonic speaks against that explanation.
More experiments involving a larger variety of data sets and exact verification methods will be needed. 
Also, there are some other limitations to our work that call for further analysis.
Firstly, we have only analysed the effect of the Lipschitz continuity induced by the prior for i.i.d. zero-mean Gaussian priors. While this is a commonly used prior \cite{wenzel2020good, wilson2020bayesian}, it would be interesting to see if similar results also hold for other parameter prior distributions (or even functional priors \cite{sun2019functional}). Secondly, we have only analysed the effect of this Lipschitz continuity on the adversarial robustness of BNNs with MFVI approximated posteriors. We are currently conducting similar analyses for BNNs with HMC approximated posteriors, which should be closer to the exact posterior for the relatively small NN architectures considered here \cite{farquhar2020liberty}.

%% file: sections/appendix.tex
\subsection{Proofs}
\subsubsection{Proof of Proposition \ref{prop:lipBNN}}

\begin{proof}
By observing that, for $k-$Lipschitz continuous activation functions, a Lipschitz constant to $f^{c}(\cdot|\theta)$ is given by $\prod_{l=1}^D \| W \|$ \citep{szegedy2013intriguing}, it can quickly be established that:
\begin{eqnarray*}
|g(x) - g(x')| & = &\left| \int{ \left(f^{c}(x|\theta) - f^{c}(x'|\theta)\right) p(\theta)}d\theta \right| \\
& \leq &\int{ \left|f^{c}(x|\theta) - f^{c}(x'|\theta)\right| p(\theta)}d\theta \\
& \leq &\int{ k^D \prod_{l=1}^D \|W^l\| \|x-x'\| p(\theta)}d\theta \\
&= & k^D \prod_{l=1}^D \mathbb{E}_{p}[\|W^l\|] \|x - x'\|
\end{eqnarray*}
\end{proof}

\subsubsection{Proof of Propositon \ref{prop:lipNormalBNN}}

\begin{proof}
We must hence show that $\mathbb{E}_p [\|W^l\|]$ is monotonic in $\alpha^2$ and $\|M^l\|$. 
We show that this holds for the Frobenius norm, as the result for other norms then follows by standard matrix norm equivalencies for both the $L_2-$ and the $L_{\infty}-$norm.
\begin{eqnarray}
    \mathbb{E}_p \left[\|W^l\|_F\right] & = & \mathbb{E}_p \left[ \sqrt{\sum_{i,j} |w^l_{i,j}|^2}\right] \\
    & \leq & \sqrt{\sum_{i,j} \mathbb{E}_p \left[(w^l_{i,j})^2\right]} \\  
    & = & \sqrt{\sum_{i,j} \left((m^l_{i,j})^2 + \alpha^2\right)} \\
    & = & \sqrt{\|M^l\|_F^2 + C\alpha^2}
\end{eqnarray}
with $C$ being the number of elements in $M^l$. The inequality follows from Jensen's inequality and the linearity of the expectation.
\end{proof}

\subsubsection{Proof of Lemma \ref{lem:priorposterior}}

\begin{proof}
Expanding out the KL term in Eq. (\ref{ELBO}), we obtain:
\begin{eqnarray} \label{eq:l2reg}
    \mathrm{ELBO}(\mathbf{m},\mathbf{s}) & = & F(\mathbf{m},\mathbf{s}) \nonumber \\
    & - & \frac{1}{2} \Bigg( 
    \log \frac{\prod_{i=1}^P \alpha^2}{\prod_{i=1}^P s_i^2} - P  \nonumber \\
    & &  + \sum_{i=1}^P \frac{s_i^2}{\alpha^2} + \frac{1}{\alpha^2} \| \mathbf{m} \|^2_2  \Bigg)
\end{eqnarray}
We thence get:
\begin{eqnarray}
\frac{\delta \mathrm{ELBO}(\mathbf{m},\mathbf{s})}{\delta s_i^2} (\alpha^2) = \frac{\delta F(\mathbf{m}, \mathbf{s})}{\delta s_i^2}  + \frac{1}{s_i^2} - \frac{1}{\alpha^2} \label{eq:dELBOds}
\end{eqnarray}
and 
\begin{eqnarray}
\frac{\delta \mathrm{ELBO}(\mathbf{m},\mathbf{s})}{\delta m_i} (\alpha^2) = \frac{\delta F(\mathbf{m}, \mathbf{s})}{\delta m_i} - \frac{m_i}{\alpha^2} \label{eq:dELBOdm}
\end{eqnarray}
Let $\bar{\mathbf{m}}(\alpha^2), \bar{\mathbf{s}}(\alpha^2)$ be the optimal solutions to maximising the ELBO for a given prior variance $\alpha^2$, i.e. the ELBO gradients in Eqs. (\ref{eq:dELBOds}) and (\ref{eq:dELBOdm}) equal $0$. 
For $\tilde{\alpha}^2 < \alpha^2$ we then get from Eq. (\ref{eq:dELBOds}) that $\frac{\delta \mathrm{ELBO}(\tilde{\mathbf{m}}(\alpha^2),\bar{\mathbf{s}}(\alpha^2))}{\delta s_i^2} (\tilde{\alpha}^2) < 0$ and thus $\bar{s_i}(\tilde{\alpha}^2) < \bar{s_i}(\alpha^2)$.\\
Analogously, from Eq. (\ref{eq:dELBOdm}) we get that if $\bar{m}_i(\alpha^2) > 0$ then $\bar{m_i}(\tilde{\alpha}^2) < \bar{m_i}(\alpha^2)$ and if $\bar{m}_i(\alpha^2) < 0$ then $\bar{m_i}(\tilde{\alpha}^2) > \bar{m_i}(\alpha^2)$, and thus in either case $\bar{m_i}^2(\tilde{\alpha}^2) < \bar{m_i}^2(\alpha^2)$ 
\end{proof}

\subsubsection{Proof of Corollary \ref{cor:MFVI}}

\begin{proof}
This follows directly from combining Lemma \ref{lem:priorposterior} with Propositions \ref{prop:lipBNN} and \ref{prop:lipNormalBNN} (the squared Frobenius norm of a mean weight matrix $M^l$ is a subset of the summands of $\|\mathbf{m}\|_2^2$). 
\end{proof}

\subsection{Details of PGD Attack on BNNs}
An attack BNNs using PGD in its original formulation was performed by drawing a new sample of parameters from the posterior at each attack iteration \cite{liu2018adv}. However, this leads to sub-optimal results as it is based on a noisy version of the mean gradient. To correct for this, a modified version which calculates the expected gradient at each PGD iteration has been suggested \cite{zimmermann2019comment}. We further improve this modified attack by two modifications. Firstly, we initialise the PGD starting point based on a random uniform draw in the perturbation ball instead of the original test point, as previous research has shown this to be more efficient for attacks on DNNs\cite{wong2020fast}. Also, other research indicates that the expected gradient might be $0$ at original test points for MFVI approximated BNNs \cite{carbone2020robustness}. Secondly, we reintroduce the originally proposed \textit{sign} function on the gradient to the attack formulation, as we find that the attacks are consistently stronger using the sign of the gradient rather than its value at each iteration (we run both versions in each experiment). 
Motivated by the fact that the expected gradient might be $0$ at or near test points \cite{carbone2020robustness}, we additionally run the original PGD attack \cite{madry2017towards} on the posterior parameter mean $\mathbf{\bar{m}}$ directly. So in total, we run 3 attacks (BNN PGD with and without \textit{sign} function + original PGD on $\mathbf{\bar{m}}$). We only report the results of the strongest attack.